\documentclass[lettersize,journal]{IEEEtran}
\usepackage{amsmath,amsfonts}
\usepackage{algorithmic}
\usepackage{algorithm}
\usepackage{array}
\usepackage[caption=false,font=normalsize,labelfont=sf,textfont=sf]{subfig}
\usepackage{textcomp}
\usepackage{stfloats}
\usepackage{url}
\usepackage{verbatim}
\usepackage{graphicx}
\usepackage{cite}
\usepackage{multirow}
\usepackage{siunitx}
\usepackage{longtable}
\usepackage{booktabs}
\usepackage{xcolor,colortbl} 
\definecolor{my_green}{RGB}{0,120,0}
\definecolor{red}{RGB}{180,0,0}
\definecolor{rowcolor1}{gray}{0.95}
\definecolor{rowcolor2}{gray}{1.0}
\definecolor{shadecolor}{gray}{0.9}
\usepackage{makecell}
\usepackage{tabularx} 
\usepackage{subfig} 
\usepackage[export]{adjustbox}

\begin{document}
	
	\title{Steering Emotional Dynamics for Art Therapy: Controllable Narrative Script Generation through Hierarchically Guided LLM Agents}
	
	\author{Suqing Wang, Qinghai Miao, Chao Guo$^*$, Yisheng Lv$^*$
		\thanks{Suqing Wang is with the Institute of Automation, Chinese Academy of Sciences, Beijing 100190, China, and also with the School of Artificial Intelligence, University of Chinese Academy of Sciences, Beijing 100049, China.}
		\thanks{Qinghai Miao is with the School of Artificial Intelligence, University of Chinese Academy of Sciences, Beijing 100049, China.}
		\thanks{Chao Guo and Yisheng Lv are with the Institute of Automation, Chinese Academy of Sciences, Beijing 100190, China.}
		\thanks{$^*$Corresponding authors.}
	}

	\IEEEpubid{Under Review}

	\maketitle
	
	\begin{abstract}
		Art therapy plays a vital role in emotional healing, in which narrative creation acts as the primary vehicle for emotional expression. Given the inherently dynamic nature of emotions during healing, narratives with finely controlled emotional fluctuations enable individuals to safely project inner conflicts and achieve emotional catharsis. Recently, with the rapid development of Large Language Models (LLMs), automated narrative generation technology has provided a new pathway to support such artistic designs. However, while existing methods can produce fluent texts, they struggle to generate narratives that adhere to specified affective trajectories, failing to meet the demands of emotion-oriented psychological healing. To address these issues, this paper proposes EC-Script, an LLM agent-based framework that enables hierarchical control of the affective trajectory in narrative generation for emotional healing. To ensure that the generated narratives strictly follow the given emotional patterns, EC-Script establishes overall narrative direction through Emotion-Trajectory Planning, propels scene-level plot development with Character-Driven Scene Generation, and regulates local emotional changes of characters via Emotion-Controlled Script Writing. Ultimately, it outputs scene-by-scene script content that remains highly consistent with the preset affective trajectory. Experimental results demonstrate that EC-Script significantly outperforms baseline methods in affective trajectory adherence, exhibiting excellent and reliable emotional controllability, thereby providing effective technical support for AI-assisted emotional healing scenarios.
	\end{abstract}
	
	
	\section{Introduction}
	
	\IEEEPARstart{A}{rt} design serves as a vital medium for emotional expression and communication. Its unique expressiveness plays an irreplaceable role in stimulating emotional resonance and promoting mental health. In the field of art therapy, narrative design acts as a pivotal method for such emotional expression\cite{laban2024sharing, amin2024wide, hasan2025empathy}. However, psychological healing unfolds as an inherently dynamic process, where participants are guided through continuous emotional shifts, experiencing sequential phases of tension, cathartic release, and cognitive recovery. By engaging with a narrative that possesses specific emotional fluctuations, individuals can safely project their inner conflicts, facilitate emotional catharsis, and ultimately achieve cognitive restructuring and psychological resolution. Thus, the core technical challenge lies not only in generating fluent text, but also in modeling and controlling the emotional trajectory of long-form narratives. Therefore, with the continuous progress in affective intelligence, exploring how to precisely steer the emotional evolution trajectory in narrative generation holds significant value for art therapy and emotional healing applications.
	
	\IEEEpubidadjcol
	
	With the rapid development of artificial intelligence, automatic narrative art generation based on LLMs has become a frontier topic in affective computing. Existing methods achieve excellent performance in textual coherence and narrative fluency, offering novel technical pathways for digital narrative creation and art design. However, these methods lack precise control capabilities in emotion progression, making it difficult to generate narrative content that strictly adheres to a preset emotional trajectory according to specific requirements. This lack of affective guidance in the generation process undermines the audience's emotional experience, thereby limiting its practical application in emotional healing scenarios.
	
	To tackle the above challenges, this paper proposes EC-Script, an LLM agent-based framework that enables hierarchical control of the emotional trajectory in narrative generation. To ensure adherence to specified emotional patterns, the framework sequentially implements Emotion-Trajectory Planning, Character-Driven Scene Generation, and Emotion-Controlled Script Writing, achieving precise control over the narrative's emotional progression. This method effectively addresses the problem of affective trajectory control, providing crucial technical support for applications such as art therapy and emotional healing.
	
	Our contributions are summarized as follows:
	
	\begin{itemize}
		\item{We propose EC-Script, an LLM agent-based framework that generates long-form scripts with controllable affective trajectories, providing foundational narrative prototypes for therapeutic applications such as theatrical performances and interactive narrative experiences.}
		\item{We design a hierarchical narrative emotional control approach. Through a cascaded architecture of global narrative planning, character-centered scene orchestration, and fine-grained local script writing, we achieve effective control over the overall narrative direction, the scene-level plot progression, and the corresponding local character emotional changes.}
		\item{Experimental results show that our method significantly outperforms all baseline models in adherence to specified emotional arcs, while maintaining competitive narrative coherence, relevance, and interestingness. Ablation experiments further validate the effectiveness of each module in emotional control.}
	\end{itemize}
	
	The structure of this paper is as follows: Section II reviews related work on LLM-based narrative generation and narrative affective modeling. Section III introduces key concepts of emotional arcs and VAD emotional parameters, and presents the task formulation. Section IV details the EC-Script framework. Section V presents the experimental results and analysis. Section VI concludes the paper.
	
	\begin{table*}[htbp!]
		\vspace{-5pt}
		\caption{The six basic emotional arc types, which reflect the overall narrative direction and emotional fluctuations\cite{reagan2016emotional, tian2024large}.}
		\centering
		\renewcommand{\arraystretch}{1.5}
		\begin{tabular}{@{} *{6}{>{\centering\arraybackslash}p{0.15\textwidth}} @{}}
			\toprule
			\textbf{Rags to Riches} &
			\textbf{Tragedy} &
			\textbf{Icarus} &
			\textbf{Man in a Hole} &
			\textbf{Cinderella} &
			\textbf{Oedipus} \\[2pt]
			
			\raisebox{-0.1\height}{\includegraphics[width=0.13\textwidth, height=0.08\textheight, keepaspectratio]{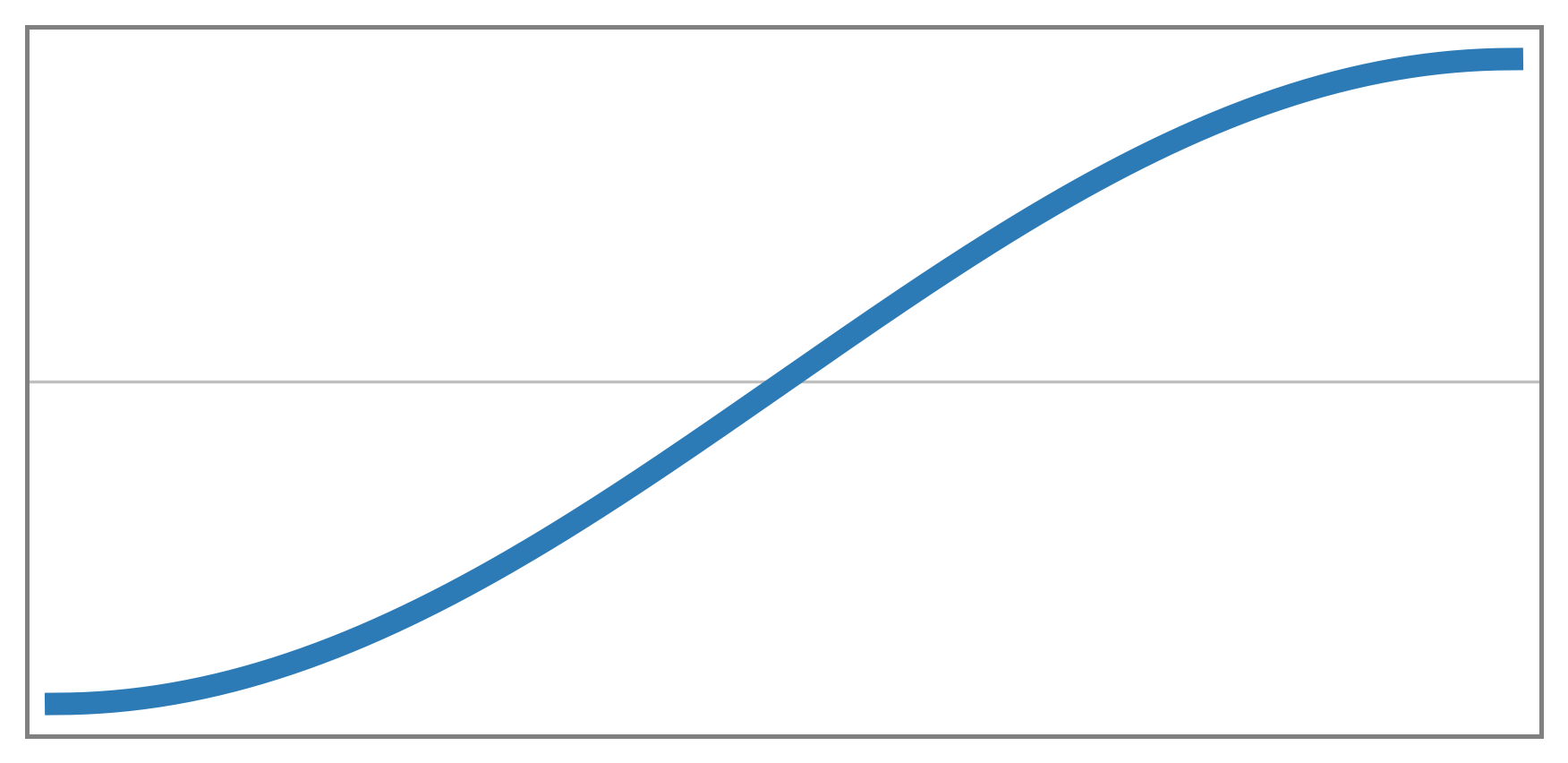}} &
			\raisebox{-0.1\height}{\includegraphics[width=0.13\textwidth, height=0.08\textheight, keepaspectratio]{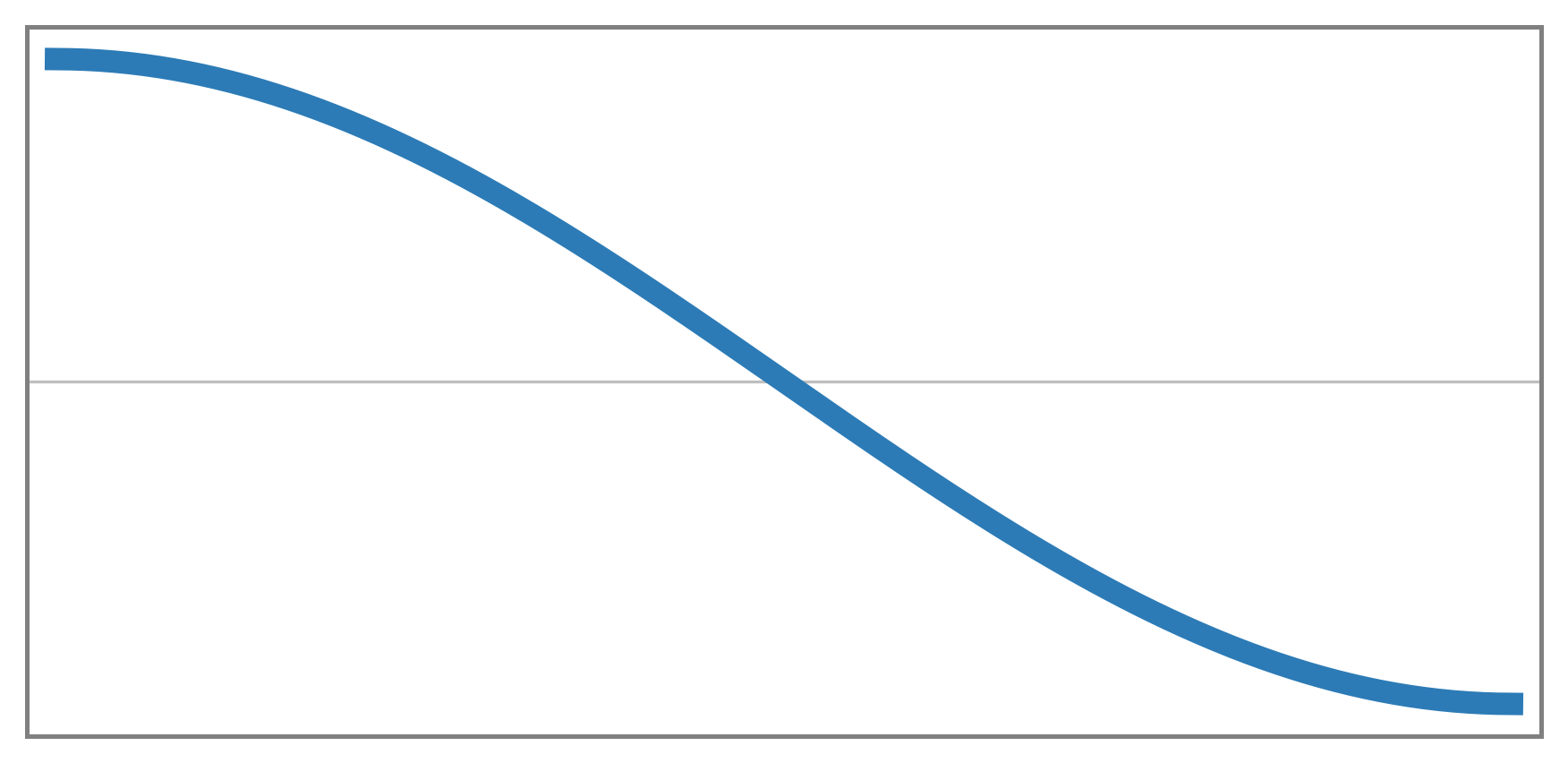}} &
			\raisebox{-0.1\height}{\includegraphics[width=0.13\textwidth, height=0.08\textheight, keepaspectratio]{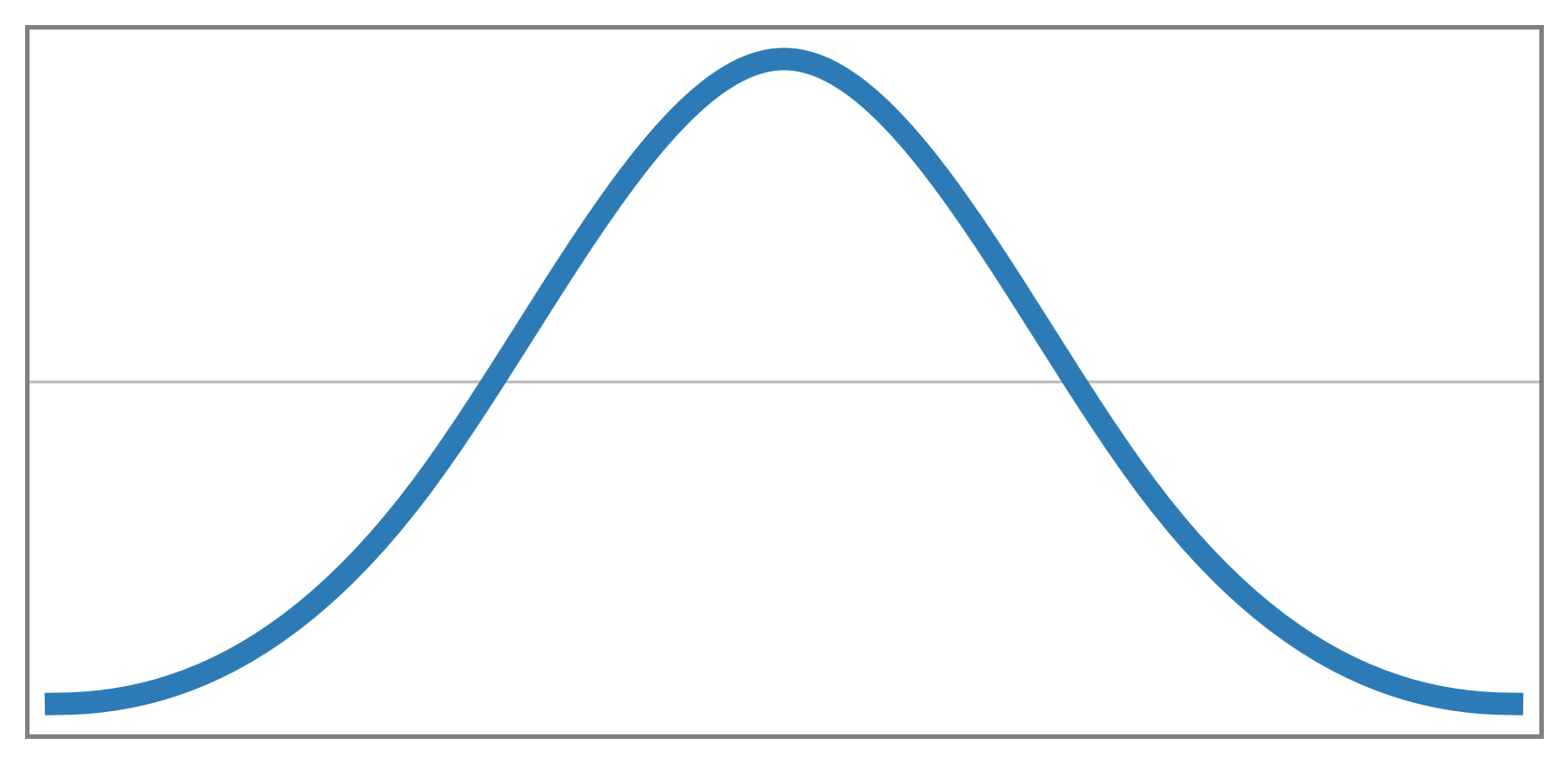}} &
			\raisebox{-0.1\height}{\includegraphics[width=0.13\textwidth, height=0.08\textheight, keepaspectratio]{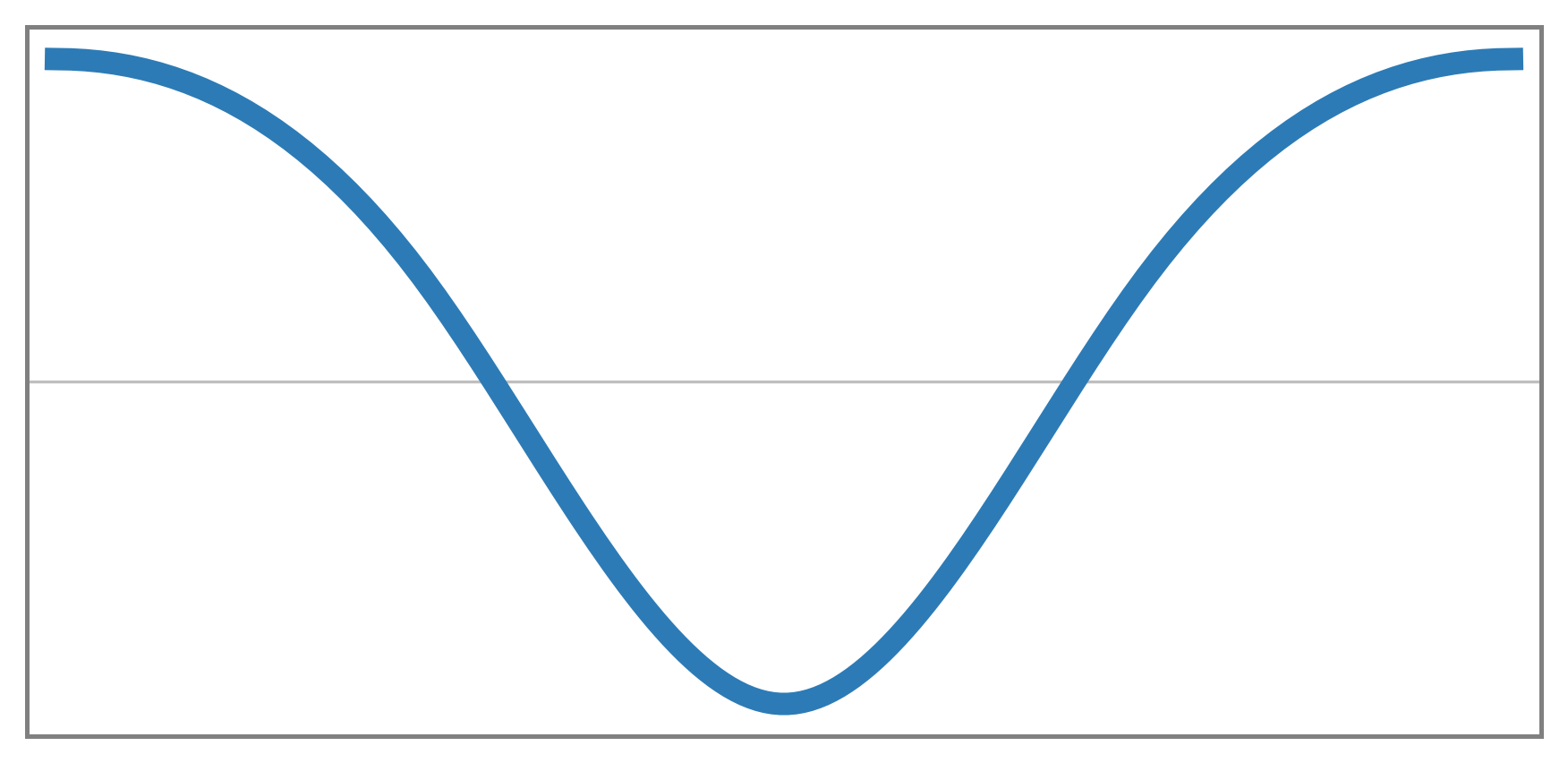}} &
			\raisebox{-0.1\height}{\includegraphics[width=0.13\textwidth, height=0.08\textheight, keepaspectratio]{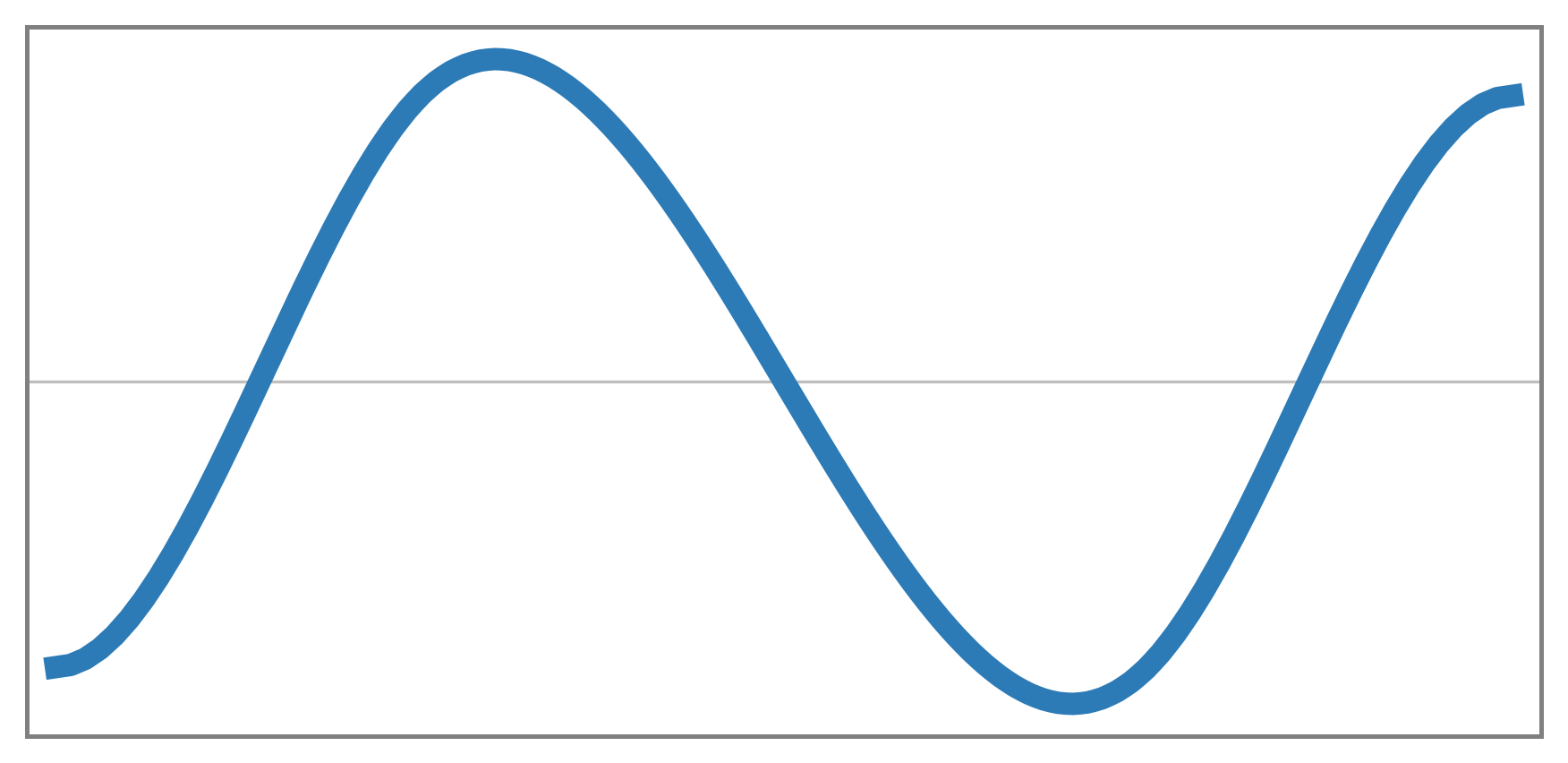}} &
			\raisebox{-0.1\height}{\includegraphics[width=0.13\textwidth, height=0.08\textheight, keepaspectratio]{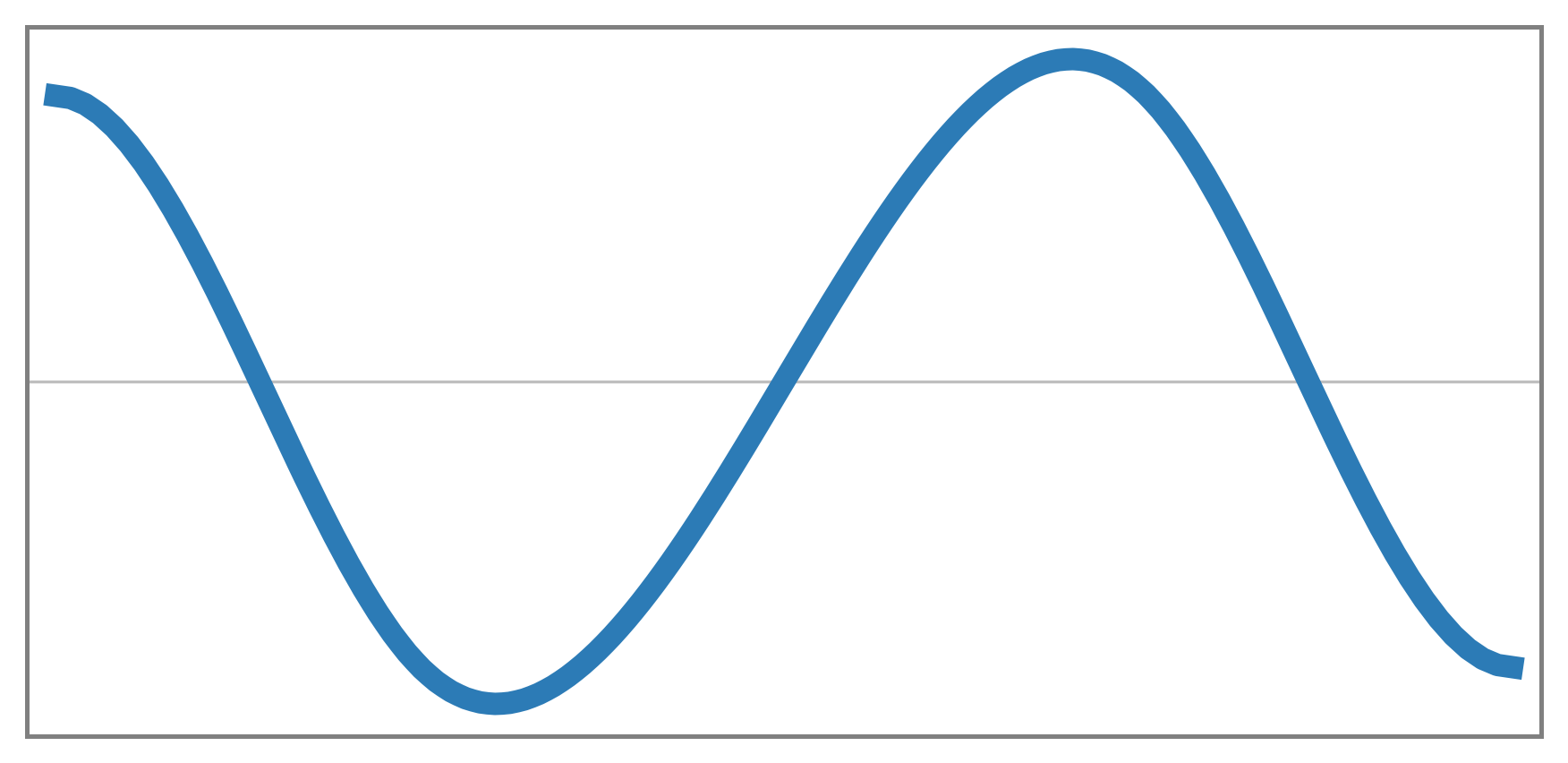}} \\[3pt]
			
			\small A continuous rise from humble beginnings to a state of lasting success.
			&
			\small A continuous fall from a position of prominence to a state of total ruin.
			&
			\small Starts low, reaches a peak or dream and finally suffers a downfall. 
			&
			\small Starts high, has a dilemma or crisis and finally finds a way out. 
			&
			\small A rise followed by a temporary setback, ending in a triumphant recovery.
			&
			\small A fall followed by a brief recovery, ending in a definitive collapse.
			\\ \addlinespace[5pt] 
			
			\footnotesize e.g. \textit{The Winter's Tale} &
			\footnotesize e.g. \textit{Lady Susan} &
			\footnotesize e.g. \textit{Shadowings} &
			\footnotesize e.g. \textit{The Magic of Oz} &
			\footnotesize e.g. \textit{The Mystery of the Hasty Arrow} &
			\footnotesize e.g. \textit{This World is Taboo} \\
			
			\bottomrule
		\end{tabular}
		\label{tab:emotional_arcs}
	\end{table*}
	
	\section{Related Work}
	\subsection{LLM-Based Narrative Generation}
	
	Automatic narrative generation has advanced rapidly in recent years, gradually shifting from deep learning-based sequence prediction to LLM-based generation paradigms. These paradigms can be mainly divided into hierarchical generation based on outline planning and collaborative generation based on multi-agent interaction.
	
	Hierarchical Generation Based on Outline Planning. Plan-and-Write\cite{yao2019plan} establishes the classic two-stage paradigm of ``plan first, then write". DOC\cite{yang2023doc} proposes a ``coarse-to-fine" fine-grained outline control mechanism and strictly aligns outline details during the generation phase to avoid plot deviation. Dramatron\cite{mirowski2023co} further concretizes the outline into independent modules such as titles, characters, scenes, and dialogues. MoPS\cite{ma2024mops} improves story generation quality by refining the premise that guides generation. LOGIC\cite{liu2025logic} continuously refines outlines by combining imitation learning and self-critique. DOME\cite{wang2025generating} introduces temporal graphs for chronological plot modeling, constructing a dynamic hierarchical outline that evolves with the plot. WritingPath\cite{lee2025navigating} explicitly constructs an executable writing path, using multi-level outlines as intermediate representations to progressively guide LLMs from high-level conceptualization to specific text grounding. STORYTELLER\cite{li2025storyteller} deconstructs outlines into plot nodes based on SVO (Subject-Verb-Object) triplets and integrates dynamic interactions between storylines and knowledge graphs to enhance story coherence and logic.
	
	Collaborative Generation Based on Multi-Agent Cooperation. Agents' Room\cite{huot2025agents} decomposes the story creation task into multiple subtasks, enhancing the quality of long-form narrative generation through multi-agent collaboration. Numerous studies have also employed multi-agent systems to construct role-playing mechanisms. IBSEN\cite{han2024ibsen} builds a ``Director-Actor" collaborative framework, which supports interactive script generation. HoLLMwood\cite{chen2024hollmwood} establishes a ``Writer-Editor-Actor" multi-level collaborative architecture and iteratively optimizes the generated results. CoDi\cite{kim2025codi} coordinates character behaviors through a reinforced director agent to achieve structurally controllable and flexible interactive narrative generation. HAMLET\cite{jiang2025hamlet}  employs a hierarchical adaptive multi-agent architecture, offline constructing a narrative blueprint, and relies on actor agents with memory, goal, and emotion attributes to accomplish real-time dramatic generation and embodied performance. DiriGent\cite{yang2025steering} introduces a dynamic belief system, utilizing cognitive conflicts between a character's ideals and reality to drive behavioral decision-making and plot evolution.
	
	Although the above methods have made significant progress in narrative generation, they still struggle to generate long narratives that conform to specified emotional patterns, thus failing to meet the needs of emotional therapy. To this end, this paper proposes EC-Script, introducing a hierarchical control strategy based on emotional arcs to achieve narrative generation with controllable affective trajectories.
	
	\subsection{Narrative Affective Modeling}
	Computational linguistics provides essential tools for modeling story structures and revealing their impact on text quality. In terms of narrative representation, Vonnegut\cite{vonnegut1999palm} introduced the concept of story shapes, describing narrative arcs through a few basic narrative elements. Reagan et al. \cite{reagan2016emotional} proposed six basic emotional arcs through automatic clustering analysis of a corpus of 1,300 books, which became the foundation for subsequent quantitative analysis of emotional arcs. Labatut and Bost\cite{labatut2019extraction} attempted to represent relationships between characters as dynamic graphs. Toubia et al.\cite{toubia2021quantifying} proposed concepts of narrative speed, volume, and circuitousness through text vector embeddings. Boyd et al.\cite{boyd2020narrative} revealed a narrative structure comprising three elements: staging, plot progression, and cognitive tension.
	
	For emotional arc modeling, methods primarily include unsupervised approaches\cite{ohman2024combining} based on lexicons (such as NRC-VAD\cite{mohammad2018obtaining}, VADER\cite{hutto2014vader}, etc.) and supervised learning approaches\cite{christ2024modeling,bizzoni2023comparing} based on deep learning. The dynamic characteristics of emotional arcs can be used to mine the evolution patterns of characters' emotions during the narrative process\cite{hipson2021emotion,vishnubhotla2024emotion}. Further research found a close correlation between the emotional arc of a story and the literary quality and reader preferences of the work. Fractal analysis reveals that the complexity and coherence of emotional arcs can serve as important indicators for evaluating novel quality\cite{hu2021dynamic,ohman2024emotionarcs}; meanwhile, the shape of specific emotional arcs can also be used to predict a book's probability of success\cite{maharjan2018letting,bizzoni2022fractal,bizzoni2023good}.
	
	Based on the emotional arc modeling method proposed by Reagan et al.\cite{reagan2016emotional}, this paper employs LLM agents to construct a hierarchical control strategy to achieve precise control over emotional narrative trajectories.
	
	\begin{figure*}[!t]
		\centering
		\includegraphics[width=\linewidth]{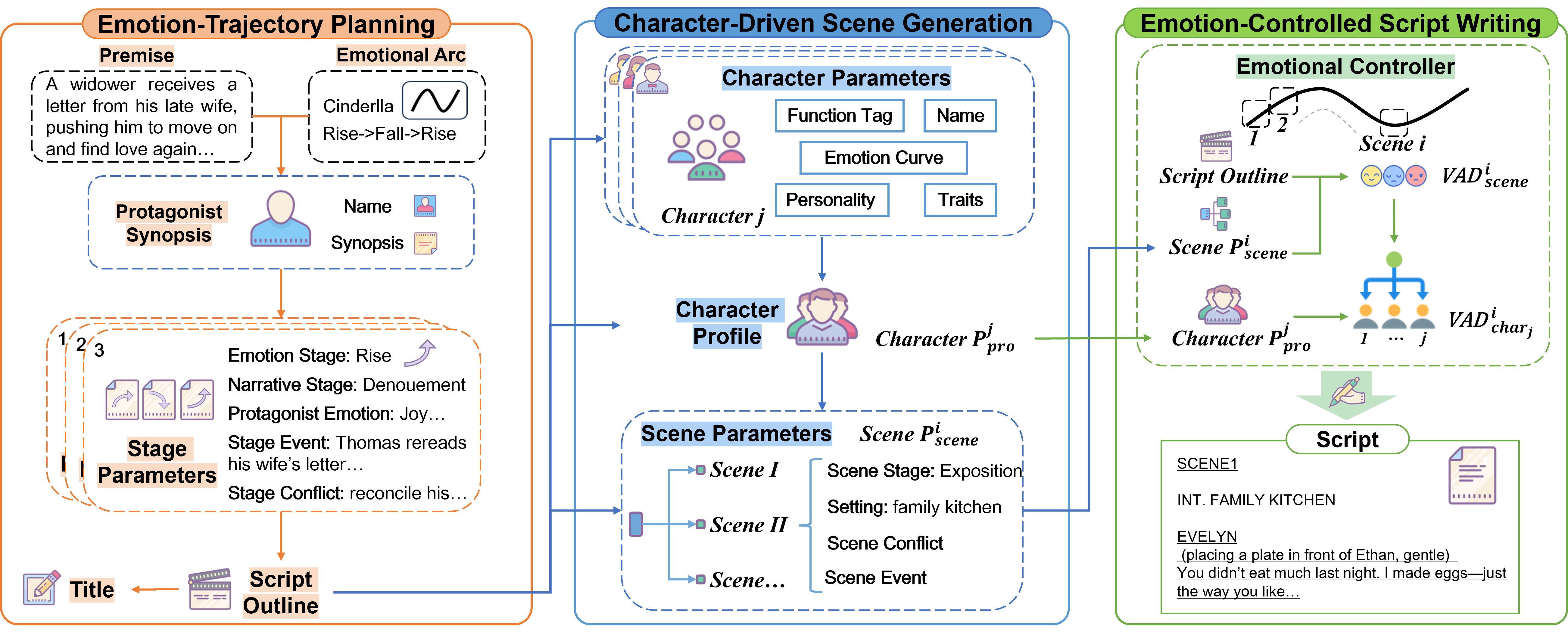} \hfill
		\caption{The overall framework of EC-Script. (1) The Emotion-Trajectory Planning module macro-plans and generates a script outline conforming to the specified emotional arc based on the user-input premise and selected emotional arc; (2) The Character-Driven Scene Generation module focuses on the coupling of characters and plots, driving the detailed planning of scene-level plots through parameterized character modeling; (3) The Emotion-Controlled Script Writing module generates fine-grained emotional control parameters for each scene and related characters, and generates the script text scene by scene to ensure the script conforms to the preset emotional trajectory.}
		\label{overall}
	\end{figure*}
	
	\section{Task Formulation}
	
	An emotional arc structurally depicts the process of a protagonist's emotional changes as the plot advances, intuitively reflecting the overall direction and emotional fluctuations of the narrative, such as a narrow escape or a hero's journey. In this paper, we adopt the emotional arc as the representation of narrative direction and use it to control the emotional development trajectory of characters during the script generation process. Following the work of Reagan et al.\cite{reagan2016emotional} and Tian et al.\cite{tian2024large}, we utilize six basic emotional arc types: Rags to Riches, Tragedy, Icarus, Man in a Hole, Cinderella, and Oedipus, as shown in Table \ref{tab:emotional_arcs}.
	
	To achieve the transition from a macro-narrative structure to micro-emotional generation, this paper further adopts VAD (Valence, Arousal, Dominance\cite{mehrabian1974approach,mohammad2018obtaining}) for fine-grained control of local plots. Valence is used to align with the overall emotional polarity direction defined by the emotional trajectory; Arousal is used to depict the change in emotional intensity as the plot advances; and Dominance is used to reflect the character's sense of emotional dominance at different stages. Compared to other modeling methods, the VAD approach can more comprehensively and accurately characterize complex local emotional changes through the synergistic interaction of Valence, Arousal, and Dominance.

	Based on the above modeling methods, the task of this paper is formulated as a controlled text generation problem. Formally, given a user-provided story premise $p$ and a target emotional arc type $arc$, our goal is to generate an emotion-controllable multi-scene script $\mathcal{S} = \{s_1, s_2, \dots, s_N\}$ of length $N$. The overall script generation process can be formulated as:
	
	\begin{equation}
		\mathcal{S} = \mathcal{F}(p, arc)
	\end{equation}
	
	where $\mathcal{F}$ represents the hierarchical controllable framework proposed in this paper. 
	
	The abstract global emotion direction $arc$ is precisely translated into local scene-by-scene textual generation. Consequently, we obtain coherent script content that adheres to the specified emotional narrative direction, providing robust technical support for AI-assisted art therapy and emotional healing.
	\newline
	\newline
	\newline
	\newline

	\section{Emotion Trajectory Guided \\ Narrative Generation}
	\subsection{Overall Framework}

	As shown in Figure \ref{overall}, EC-Script consists of three interconnected modules. Starting with the user-input premise $p$ and emotional arc $arc$, it progressively achieves hierarchical control over the narrative direction through global outline planning, character-driven scene-level plot generation, and local fine-grained control, ultimately outputting the final script $\mathcal{S}$. The complete algorithm flow is shown in Algorithm \ref{alg:et}. 
	
	\begin{algorithm}[t]
		\caption{\textsc{EC-Script}: Emotion Controllable Script Generation.}
		\label{alg:et}
		\begin{algorithmic}[1]
			\STATE \textbf{Input:} premise $p$; emotional arc $arc$
			\STATE \textbf{Output:} Final script $\mathcal{S}$
			
			\STATE \textcolor{blue}{\small\textbf{// Emotion-Trajectory Planning Module}}
			\STATE $P_{synopsis} \gets \textsc{GenProtagonistSynopsis}(p, arc)$
			\STATE $P_{stage} \gets \textsc{GenStageParams}(P_{synopsis}, p, arc)$
			\STATE $o \gets \textsc{GenScriptOutline}(P_{stage}, p, arc)$
			\STATE $title \gets \textsc{GenTitle}(o, p, arc)$
			
			\STATE \textcolor{blue}{\small\textbf{// Character-Driven Scene Generation Module}}
			\STATE $P_{char}$, $M \gets \textsc{GenCharacterParams}(o, p, arc)$
			\STATE $P_{pro} \gets \textsc{GenCharacterProfile}(P_{char})$
			\STATE $P_{scene}$, $N \gets \textsc{GenSceneParams}(o, P_{pro}, p, arc)$
			
			\STATE \textcolor{blue}{\small\textbf{// Emotion-Controlled Script Writing Module}}
			\FOR{$i \gets 1$ \TO $N$}
			\STATE $VAD_{scene}^{i} \gets \textsc{GenSceneVAD}(o, P_{scene}^i, arc)$
			\FOR{$j \gets 1$ \TO $M$}
			\STATE $VAD_{char_j}^{i} \gets \textsc{GenCharVAD}(o, P_{pro}^j, VAD_{scene}^{i})$
			\ENDFOR
			\ENDFOR
			
			\STATE \textcolor{blue}{\small\textbf{// Generate Final Script}}
			\FOR{$i \gets 1$ \TO $N$}
			\STATE $\text{SceneScript}_i \gets \textsc{GenSceneText}(
			o, \, P_{scene}^i,\, P_{pro},$
			\STATE \hspace{1.2em} $VAD_{scene}^{i}, \, VAD_{char}^{i})$
			\ENDFOR
			\STATE $\mathcal{S} \gets \textsc{MergeScript}(title, \{\text{SceneScript}_i\}_{i=1}^{N})$
			
			\RETURN $\mathcal{S}$
		\end{algorithmic}
	\end{algorithm}

	\begin{enumerate}
		\item{The Emotion-Trajectory Planning module first generates a Protagonist Synopsis describing the protagonist's development, and further constructs emotional narrative stages along with corresponding Stage Parameters. Based on this, it generates a Script Outline that delineates the overall narrative direction, providing macroscopic narrative constraints for subsequent modules.}
		\item{The Character-Driven Scene Generation module starts from character modeling to generate Character Parameters and exhaustive Character Profiles that describe characters across multiple dimensions. Then, centering on the characters, it plans the Scene Parameters for each scene based on the Script Outline, realizing the transition from narrative planning to character actions.}
		\item{The Emotion-Controlled Script Writing module utilizes the structured information generated previously to produce VAD emotion parameters for each scene and related characters. Under the constraints of these parameters, it writes the script text scene by scene, finely controlling the emotional direction of each scene, ultimately forming a complete narrative content that conforms to the target emotional trajectory.}
	\end{enumerate}
	
	As summarized in Table \ref{tab:elements_summary}, EC-Script employs a set of hierarchical representations to explicitly model the narrative from the global narrative trajectory down to the scene-level plot progression and local emotional dynamics. By structurally decomposing and controlling the emotional evolution across these multiple levels, the framework ensures that the generated narrative strictly adheres to the preset affective pattern, thereby providing a reliable and precise storytelling mechanism for art therapy and emotional healing.
	
	\subsection{Emotion-Trajectory Planning}
	
	\begin{table}[t]
		\centering
		\caption{Hierarchical representations of affective modeling.}
		\label{tab:elements_summary}
		\renewcommand{\tabularxcolumn}[1]{m{#1}}
		
		\begin{tabularx}{\linewidth}{
				>{\centering\arraybackslash}m{1.8cm}
				>{\centering\arraybackslash}X
				>{\centering\arraybackslash}m{2.5cm}
			}
			\toprule
			\textbf{Representation} & \textbf{Components} & \textbf{Function} \\ \midrule
			
			Stage Parameters & Emotion Stage, Narrative Stage, Protagonist Emotion, Stage Event, Stage Conflict & Global affective direction \\ \midrule
			
			Character Parameters & Function Tag, Traits, Personality, Emotion Curve & Character-level affective development \\ \midrule
			
			Scene Parameters & Scene Stage, Setting, Scene Conflict, Scene Event & Scene-level affective grounding \\ \midrule
			
			Scene VAD & Valence, Arousal, Dominance & Local affective environment \\ \midrule
			
			Character VAD & Transitional Valence, Arousal, Dominance & Local affective dynamics \\ \bottomrule
		\end{tabularx}
	\end{table}
	
	This module aims to conduct macroscopic narrative planning, mapping abstract emotional arcs into specific narrative stages, thereby generating a global narrative trajectory to provide a globally consistent plot skeleton for subsequent generation. Taking the Premise $p$ and Emotional Arc $arc$ as inputs, this module sequentially generates the Protagonist Synopsis $P_{synopsis}$ and Stage Parameters $P_{stage}$, ultimately outputting the Script Outline $o$ and script Title $title$.
	
	Specifically, it first generates $P_{synopsis}$ based on $p$ and $arc$. This parameter includes the protagonist's name and a synopsis describing their core experiences, thereby establishing the narrative center. $P_{stage}$ is further generated based on $P_{synopsis}$, including: the parameter Emotion Stage, which defines what phase of the arc the current emotion is in, represented by primitives such as fall, rise, peak, stable, and fluctuation; the parameter Narrative Stage, which describes the positioning of this stage in a classical narrative structure, including elements like Exposition and Rising Action; the parameter Protagonist Emotion, which defines the core emotional tone of the protagonist at this stage, consisting of Plutchik's eight basic emotions and their mixed derivative emotions; the parameter Stage Event, which is a landmark event driving the development of this stage; and the parameter Stage Conflict, which is the core internal or external developmental conflict of the characters at this stage. The number of Narrative Stages is dynamically divided according to the topological structure of $arc$. For example, when $arc$ is Cinderella (Rise-Fall-Rise), the number of emotional stages is three. To ensure that the global narrative strictly adheres to the preset emotional development pattern, a structural verification mechanism is introduced. The model extracts the parameters Protagonist Emotion from the generated $P_{stage}$ and maps them to sentiment scores via the VADER lexicon to construct an emotional sequence for comparison against the topological trend of the target $arc$. If the resulting sequence of emotional states deviates from the target pattern, a revision prompt is triggered to iteratively refine $P_{stage}$ until the emotional fluctuation sequence aligns well with the target emotional arc, laying a solid structural foundation for subsequent generation. 
	
	Finally, by integrating $p$, $arc$, and the structured information generated above, the module generates the Script Outline $o$, which defines the narrative context conforming to the target emotional pattern at the outline level and serves as a global constraint for subsequent modules. This process is formalized as:
	\begin{equation}
		o = \textsc{GenScriptOutline}(P_{stage}, p, arc)
	\end{equation}
	
\begin{figure}[t]
		\centering
		\includegraphics[width=\linewidth]{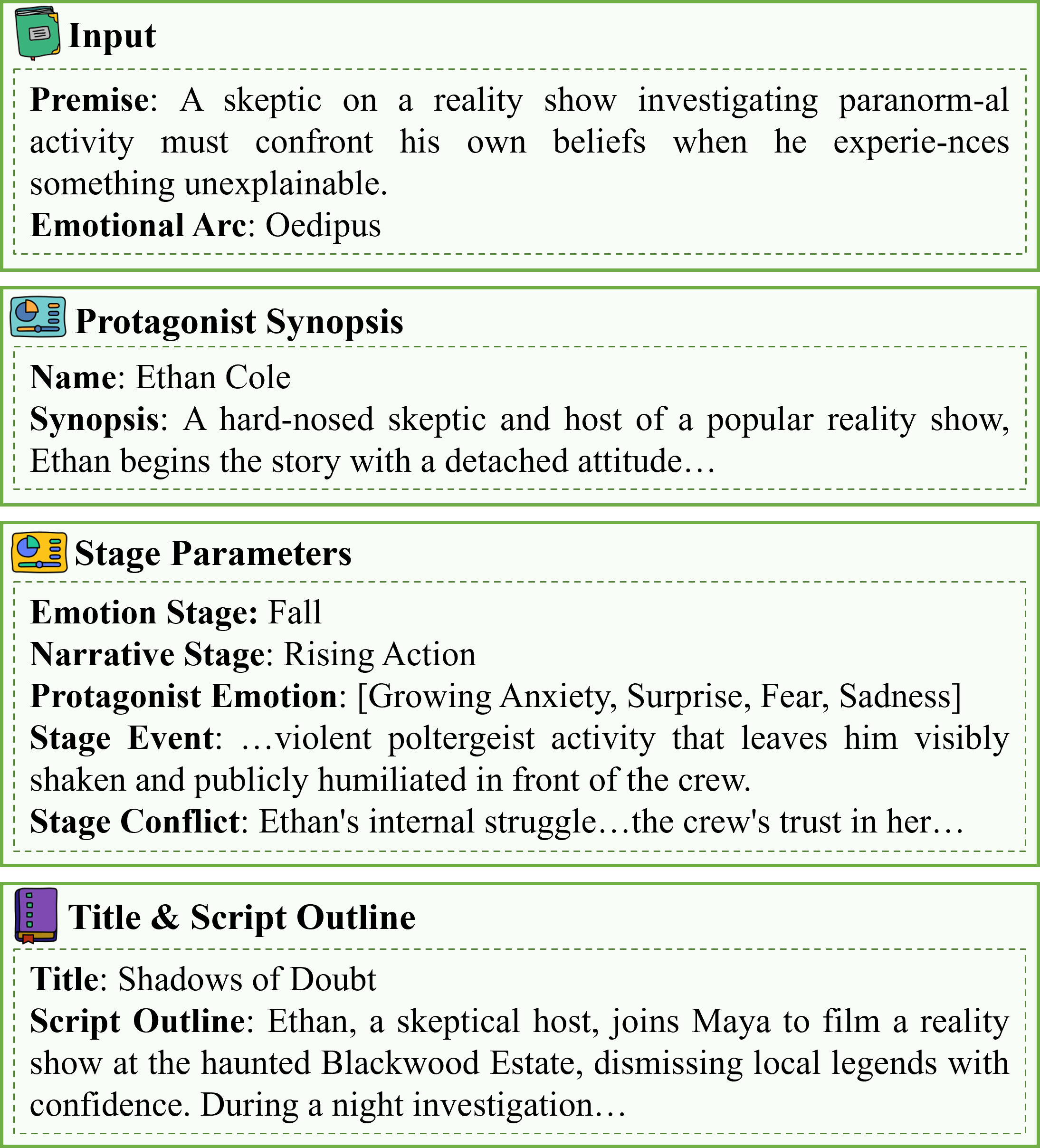}
		\caption{A generation example of Emotion-Trajectory Planning.}
		\label{module1}
	\end{figure}

	Simultaneously, an appropriate script title $title$ is generated based on $o$. Figure \ref{module1} shows an example output of this module.
	
	\subsection{Character-Driven Scene Generation}
	
	\begin{figure}[t]
		\centering
		\includegraphics[width=\linewidth]{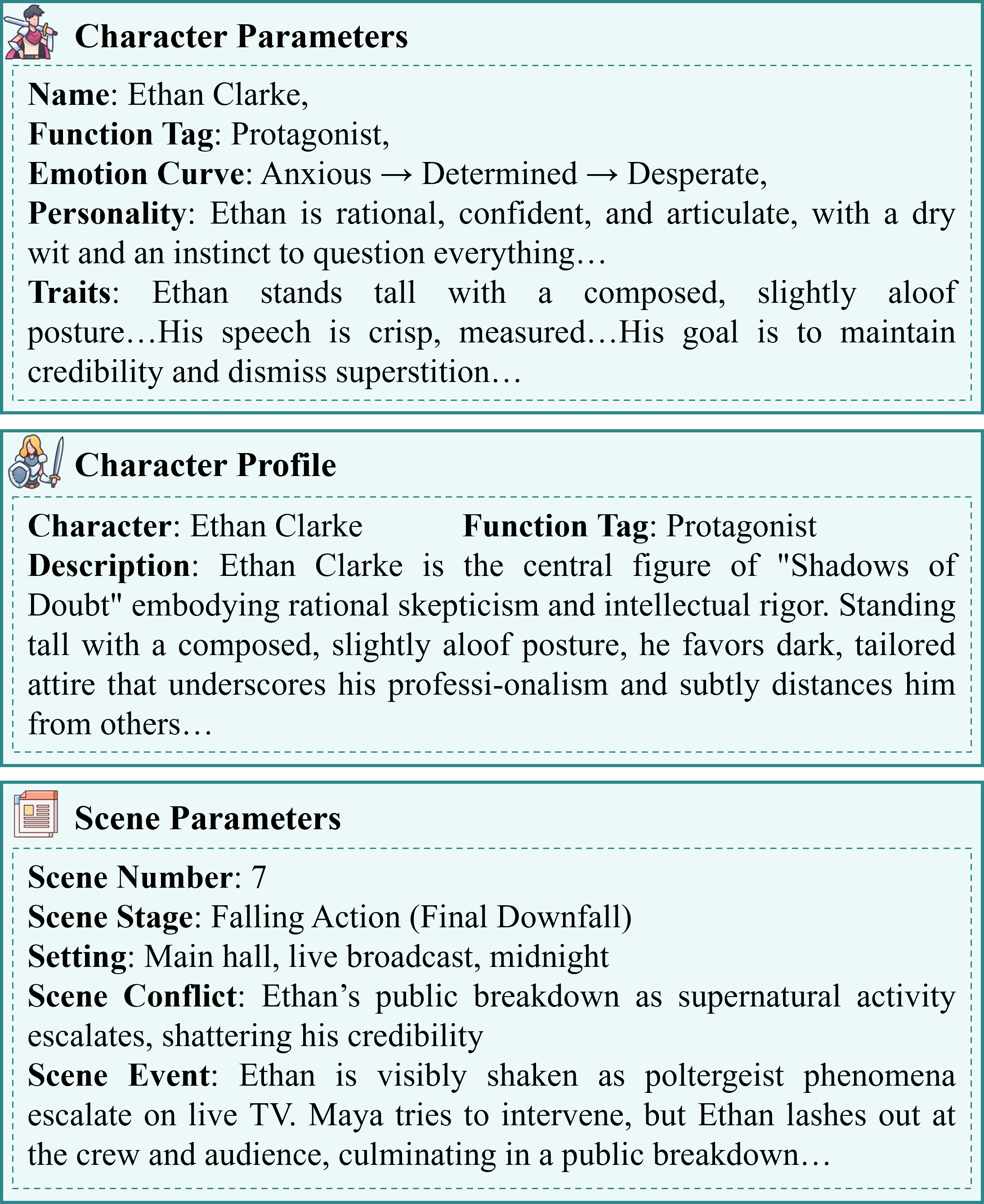}
		\caption{A generation example of Character-Driven Scene Generation.}
		\label{module2}
	\end{figure}
	
	After acquiring the global outline $o$, the granularity of the narrative needs to move from the macroscopic level to a more specific scene-by-scene level. Since an emotional arc inherently depicts the evolution of the protagonist's inner state in response to external encounters, character development and plot advancement must be highly coupled to jointly constitute the narrative of each scene. This module adopts a ``character-centric" driving mechanism, receiving $o$ generated by the preceding module along with the initial $p$ and $arc$ as inputs. It designs characters multidimensionally and accordingly drives detailed scene-level plot planning, ensuring that character development, plot settings, and the target emotional trajectory remain consistent. This module outputs Character Profiles $P_{pro}$ and Scene Parameters $P_{scene}$, which are used to guide subsequent generation.
	
	Taking $p$, $arc$, and $o$ as inputs, multidimensional Character Parameters $P_{char}$ are first generated, with the total number of characters $M$ automatically determined by the LLM. The parameters for each character include: character Name, a Function Tag distinguishing between protagonists and supporting characters, an Emotion Curve describing the character's emotional changes, and the character's internal Personality and external Traits, thereby making the character vivid and three-dimensional and ensuring their development remains consistent with the target emotional trajectory. This process is formalized as:
	\begin{equation}
		P_{char}, M = \textsc{GenCharacterParams}(o, p, arc)
	\end{equation}
	
	Based on $P_{char}$, an exhaustive Character Profile $P_{pro}$ is further generated for each character, clarifying the character's emotional development, internal motivations, and behavioral logic in the form of a complete text block.
	
	After character modeling is complete, the module slices $o$ into $N$ independent acts. For each scene $i \in [1, N]$, the model generates corresponding Scene Parameters $P_{scene}^i$, including the scene's functional Scene Stage in the overall narrative structure, the spatio-temporal Setting, the key Scene Conflict pushing the plot forward, and the specific plot design Scene Event for that scene.This process is formalized as:
	\begin{equation}
		P_{scene}, N = \textsc{GenSceneParams}(o, P_{pro}, p, arc)
	\end{equation}
	
	By aligning character attributes with affective trajectories and using this to drive the plot forward, the Character-Driven Scene Generation module achieves the transition from global narrative to specific scenes, ensuring that the plot of each scene not only fits character personalities but is also highly consistent with the global emotional trajectory. Figure \ref{module2} displays a generation example of this module.
	
	\subsection{Emotion-Controlled Script Writing}
	
	\begin{figure}[t]
		\centering
		\includegraphics[width=0.9\linewidth]{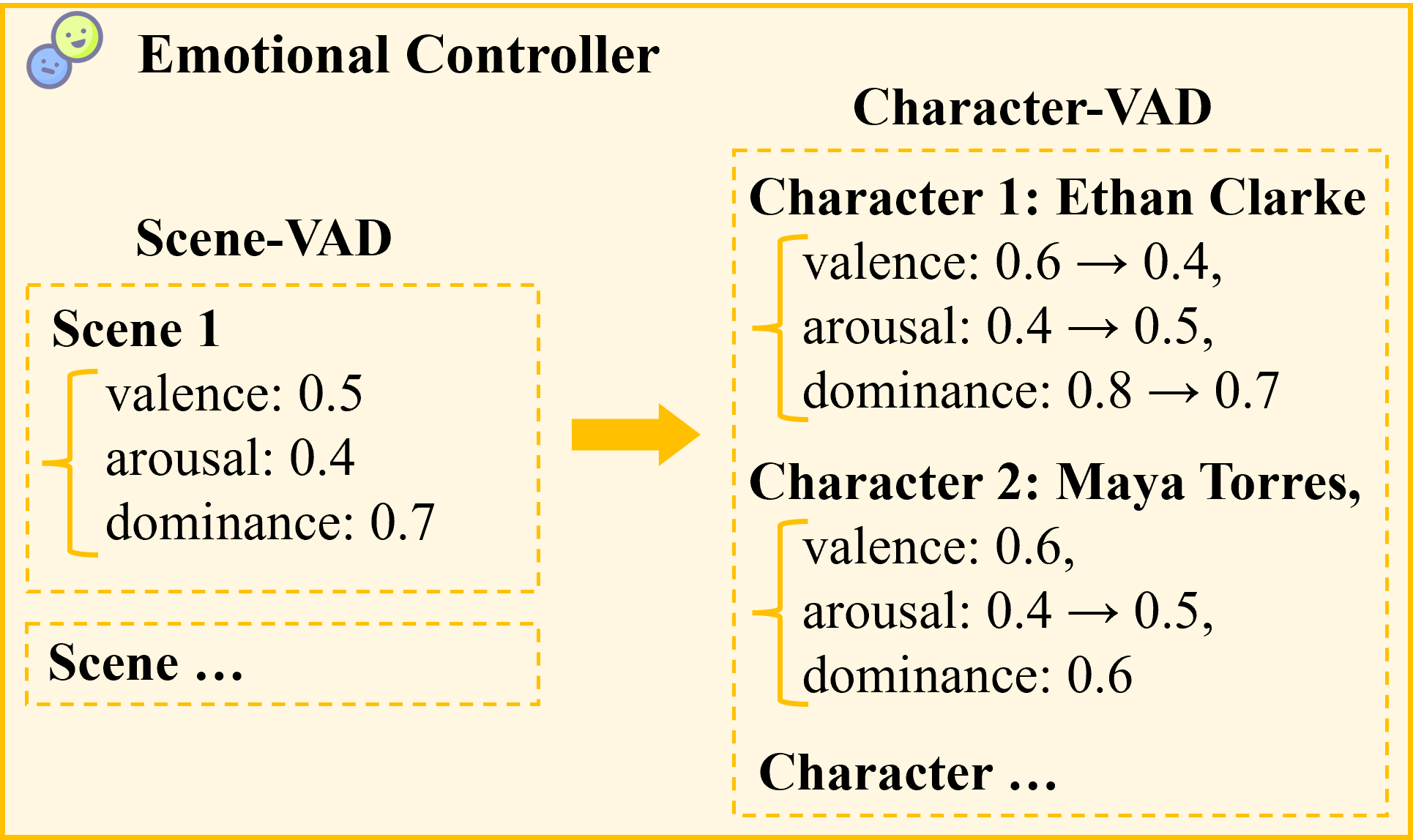}
		\caption{A generation example of the VAD parameter in Emotion-Controlled Script Writing.}
		\label{module3}
	\end{figure}
	
	To achieve fine-grained control over the expressed emotions at the script text level, this module receives $o$, $P_{pro}$, and $P_{scene}^i$ as inputs. By generating three-dimensional VAD (Valence, Arousal, Dominance) emotional parameters, it explicitly constrains textual details of every scene, thereby completing the scene-by-scene script writing.
	
	First, based on the Script Outline $o$ output by the aforementioned modules, the Scene Parameters $P_{scene}^i$ for each scene $i$, and the target emotional arc $arc$, the model generates scene-level VAD parameters $VAD_{scene}^{i} \in [0.0, 1.0]^3$ to define the overall emotional tone of the scene (encompassing emotional polarity, emotional intensity, and sense of control). This process is formalized as:
	\begin{equation}
		VAD_{scene}^{i} = \textsc{GenSceneVAD}(o, P_{scene}^i, arc)
	\end{equation}
	
	To ensure that various characters possess emotional expressions conforming to their own settings within the same scene, the model further integrates the specific Character Profile $P_{pro}^j$ to generate character-level emotional parameters $VAD_{char_j}^{i} \in [0.0, 1.0]^3$ for each character $j \in [1, M]$ in the current scene $i$:
	\begin{equation}
		VAD_{char_j}^{i} = \textsc{GenCharVAD}(o, P_{pro}^j, VAD_{scene}^{i})
	\end{equation}
	
	To accurately capture intra-scene emotional variations, these parameters are modeled in two forms: a single value indicating a constant emotional state throughout the scene, or a transitional pair (i.e., $V_{start} \rightarrow V_{end}$) denoting a dynamic emotion shift driven by character interactions. This precisely depicts their emotional states and detailed psychological evolutions in specific contexts.
	
	Finally, under the global guidance of $o$ and the local constraints of multi-level emotional parameters including $P_{pro}$, $P_{scene}^i$, $VAD_{scene}^{i}$, and the collective character VAD parameters for the scene (denoted as $VAD_{char}^{i}$), the Emotion-Controlled Script Writing module generates the script text scene by scene. The generation of the script text $\text{SceneScript}_i$ for each scene is executed as follows:
	\begin{equation}
		\begin{split}
			\text{SceneScript}_i = \textsc{GenSceneText}(o,\, P_{scene}^i,\, P_{pro}, \\
			VAD_{scene}^{i},\, VAD_{char}^{i})
		\end{split}
	\end{equation}
	
	These individual scene scripts $\{\text{SceneScript}_i\}_{i=1}^{N}$ are eventually merged alongside the $title$ into a complete script $\mathcal{S}$ conforming to the specified emotional narrative trajectory. Figure \ref{module3} shows an example of VAD emotional parameter generation during this module's process.
	
	\section{Experimental Results}
	\subsection{Experimental Design}
	\begin{figure*}[t]
		\centering
		\includegraphics[width=\linewidth]{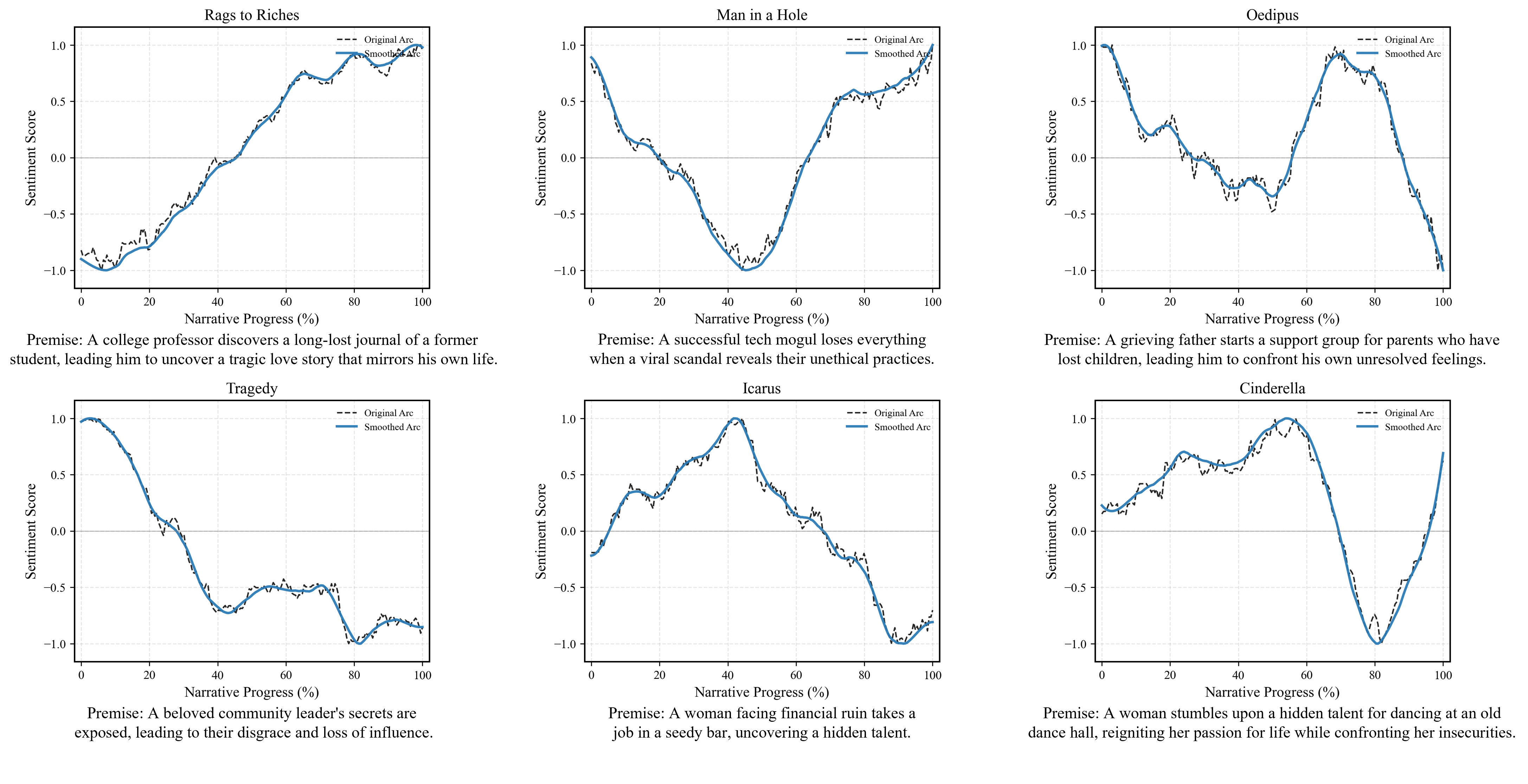}
		\caption{Examples of emotional arcs extracted from generated scripts across six different emotional arc types. The horizontal axis represents narrative progress, and the vertical axis represents the sentiment score. Results show that the emotional arcs extracted from generated scripts are highly consistent with the target standard emotional arcs.}
		\label{examples_arcs}
	\end{figure*}
	
	\subsubsection{Model Settings}
	In our experiments, we selected GPT-4.1\cite{openai2025gpt41}, Qwen3-235B-A22B-Instruct\cite{qwen3technicalreport}, and Claude-Sonnet-4.5\cite{anthropic2025sonnet45} as the backbone models for the generation phase, while employing Gemini-2.5-Pro\cite{comanici2025gemini} as the evaluation model. By cross-validating different series of models, we eliminated potential self-preference bias inherent in specific models, ensuring the objectivity and generality of the evaluation results. The LLMs used for generation were uniformly set with a temperature of 0.7 and a top-p of 0.95. The evaluation model's temperature was set to 0 to improve the stability and reproducibility of the evaluation results.
	
	\subsubsection{Dataset}
	Considering the costs of long-text script generation and test dataset collection, current related methods generally adopt a testing data scale of 20--80. Following the settings of previous work, we constructed a dataset containing 48 test samples, covering seven themes: romance, fantasy, mystery, adventure, science fiction, growth, and horror. Each sample consists of a premise and a specified emotional arc type. During construction, we uniformly and randomly sampled the six basic emotional arcs and assigned a specific emotional arc type to each premise to ensure an even distribution across categories, avoiding the impact of emotional type bias on the experimental results.
	
	\subsection{Baseline Methods}
	We selected three representative methods in the fields of script and story generation as baselines. To ensure a fair comparison, appropriate adjustments were made to these methods so that they produce structurally consistent script texts. Specifically, the baselines include:
	
	\begin{itemize}
		\item{\textbf{LLM-Plan-and-Write}: Directly plans and generates scripts using an LLM;}
		\item{\textbf{Dramatron}\cite{mirowski2023co}: A classic hierarchical generation framework designed for narrative creation;}
		\item{\textbf{Agents' Room}\cite{huot2025agents}:A multi-agent collaborative framework.}
	\end{itemize}
	
	All baselines were evaluated under identical experimental settings, length constraints, and input data (premise, specified emotional arc type, and its textual description).
	
	\subsection{Results}
	
	\begin{table*}[t]
		\caption{Performance Comparison. Our method shows clear superiority in emotional controllability (Alignment, LLM evaluation metric; DTW, objective metric), while maintaining competitive performance in Coherence, Relevance, and Interestingness.}
		\centering
		\renewcommand{\arraystretch}{1.2}
		\resizebox{\linewidth}{!}{
			\begin{tabular}{@{}ccccc >{\columncolor{gray!18}}c >{\columncolor{gray!18}}c c c @{}}
				\toprule
				\multirow{2}{*}[-3.5pt]{\textbf{Backbone}} & \multirow{2}{*}[-3.5pt]{\textbf{Model}}
				& \multicolumn{4}{c}{\textbf{LLM Metrics}} & \multicolumn{3}{c}{\textbf{Objective Metrics}} \\
				\cmidrule(lr){3-6} \cmidrule(lr){7-9}
				&
				& \textbf{Coherence$\uparrow$}
				& \textbf{Relevance$\uparrow$}
				& \textbf{Interestingness$\uparrow$}
				& \textbf{Alignment$\uparrow$}
				& \textbf{DTW$\downarrow$}
				& \textbf{DistinctL-4$\uparrow$}
				& \textbf{Length} \\
				\midrule
				
				\multirow{4}{*}{GPT-4.1}
				& LLM-Plan-and-Write		 & 57.90 $\pm$ 13.45 & 86.12 $\pm$ 13.56 & 45.95 $\pm$ 14.31 & 72.01 $\pm$ 25.44 & 62.32 & 11.61 & 1708.96 \\
				& Dramatron    & 43.99 $\pm$ 18.46 & 81.90 $\pm$ 14.55 & 42.65 $\pm$ 17.78 & 63.62 $\pm$ 31.62 & 66.76 & 11.42 & 3340.54 \\
				& Agents' Room & \textbf{74.03}* $\pm$ 13.17 & \textbf{93.78}* $\pm$ 2.84 & 61.43* $\pm$ 9.34 & 75.49 $\pm$ 27.63 & 65.84 & 12.23 & 3129.12 \\
				& Ours         & 68.72* $\pm$ 20.67 & 89.97* $\pm$ 14.67 & \textbf{62.53}* $\pm$ 10.36 & \textcolor{red}{\textbf{91.12}}* $\pm$ 5.76 & \textcolor{red}{\textbf{46.52}} & \textbf{12.37} & 5944.50 \\
				
				\midrule
				
				\multirow{4}{*}{\makecell{Qwen3-235B\\-A22B-Instruct}}
				& LLM-Plan-and-Write          & 68.45 $\pm$ 12.46 & 93.29* $\pm$ 4.76 & 58.50 $\pm$ 10.15 & 78.64 $\pm$ 23.40 & 63.27 & 12.31 & 2792.96 \\
				& Dramatron    & 43.83 $\pm$ 22.52 & 75.11 $\pm$ 25.95 & 42.26 $\pm$ 17.14 & 70.21 $\pm$ 28.43 & 63.89 & 9.80 & 3987.00 \\
				& Agents' Room & \textbf{85.95}* $\pm$ 10.02 & \textbf{93.60}* $\pm$ 3.93 & \textbf{83.23}* $\pm$ 5.70 & 81.90 $\pm$ 23.44 & 67.94 & 12.63 & 4476.08 \\
				& Ours         & 77.69 $\pm$ 16.31 & 90.48* $\pm$ 15.07 & 79.49 $\pm$ 8.79 & \textcolor{red}{\textbf{92.38}}* $\pm$ 8.45 & \textcolor{red}{\textbf{57.96}} & \textbf{12.66} & 7057.23 \\
				
				\midrule
				
				\multirow{4}{*}{\makecell{Claude\\-Sonnet-4.5}}
				& LLM-Plan-and-Write          & 71.36 $\pm$ 11.35 & 90.44 $\pm$ 11.42 & 58.49 $\pm$ 11.15 & 80.67 $\pm$ 24.77 & 71.31 & 12.30 & 3337.06 \\
				& Dramatron    & 32.76 $\pm$ 16.51 & 72.61 $\pm$ 22.42 & 33.88 $\pm$ 12.86 & 73.08 $\pm$ 25.45 & 75.73 & 10.58 & 7111.98 \\
				& Agents' Room & \textbf{87.64}* $\pm$ 7.59 & \textbf{92.25}* $\pm$ 2.04 & 78.34* $\pm$ 7.95 & 85.00 $\pm$ 21.20 & 64.21 & 12.54 & 4118.62 \\
				& Ours         & 82.90* $\pm$ 14.46 & 91.50* $\pm$ 12.99 & \textbf{79.31}* $\pm$ 9.33 & \textcolor{red}{\textbf{93.82}}* $\pm$ 8.05 & \textcolor{red}{\textbf{49.36}} & \textbf{12.78} & 7968.21 \\
				\bottomrule
		\end{tabular}}
		\begin{flushleft}
			\footnotesize
			Bold indicates the best results, and * denotes statistical significance ($p < 0.05$). 
			We highlight \setlength{\fboxsep}{1.5pt}\colorbox{gray!18}{\textbf{Alignment}} and \colorbox{gray!18}{\textbf{DTW}} 
			as primary metrics for evaluating emotional trajectory adherence.
		\end{flushleft}
		\label{tab:main_results}
	\end{table*}
	
	\subsubsection{Evaluation of Emotional Arc Adherence}
	
	Since evaluating emotional adherence is our primary objective, we extracted the emotional arcs of the generated scripts to measure how strictly they follow the given narrative patterns and affective trajectories. Alongside visual comparisons with standard emotional arcs, we performed quantitative evaluations by measuring the similarity of the emotion sequences.
	
	While recent research has shown that LLMs and traditional dictionary-based methods perform comparably in emotional arc extraction tasks \cite{teodorescu2023evaluating}, we opted for the latter to ensure stability and interpretability. Specifically, we followed the methodology of \cite{reagan2016emotional} to extract emotional arcs: we assigned word-level emotional values to the text based on the LabMT emotion lexicon\cite{dodds2011temporal} and used a sliding window mechanism for segmented statistical analysis, thereby obtaining an emotional time series that varies with narrative progress. Subsequently, a Savitzky-Golay filter was used to smooth the extracted raw sequences to reduce local noise interference. The sequences were then resampled to a fixed length through interpolation. Finally, Min-Max normalization was employed to scale the emotion values into the $[-1, 1]$ interval, thereby eliminating the impact of scale differences on subsequent calculations.
	
	Figure \ref{examples_arcs} shows examples of emotional arcs extracted from EC-Script generation results, where the gray curve is the raw emotional curve and the blue curve is the smoothed emotional curve. Visualization results indicate that the extracted emotional arcs highly match the standard emotional arcs in overall trend, validating the effectiveness of our method in generating narrative content consistent with specified emotional patterns.
	
	To further quantitatively evaluate our method's advantage in tracking affective trajectories, we employed Dynamic Time Warping (DTW\cite{sakoe2003dynamic}) to calculate the distance between the generated emotional arcs and the target patterns, comparing these with baseline methods. DTW is a classical method for comparing the similarity of time series, capable of robustly calculating the overall shape similarity between extracted and target emotional arcs. Referencing the aforementioned studies, we generated six standard emotional arcs as target patterns using sine functions and cubic spline interpolation to characterize typical emotional evolution trends. For each method, we extracted the emotional arcs of all generated scripts according to the above steps, calculated their DTW similarity with the specified emotional patterns, and measured the adherence of the narrative content to the specified emotional pattern by calculating the average DTW score.
	
	The similarity results between the generated narrative content and specified emotional arcs are shown in the DTW column of Table \ref{tab:main_results}. A smaller distance indicates that the emotional arc is closer to the target pattern; a larger distance indicates a higher degree of deviation. Under three different backbone settings, EC-Script significantly outperformed all baseline methods on the DTW metric, demonstrating from an objective measurement perspective that this method can effectively constrain generated texts to follow specified emotional development patterns.
	
	\subsubsection{LLM-Based Overall Evaluation}
	
	Current LLMs possess strong contextual understanding capabilities, can simulate human reading experiences, and are widely used in automated evaluations of story generation\cite{chhun2024language}. Following this practice, we chose to use an LLM to evaluate the generated scripts across the following four quality dimensions using a Likert scale (out of 100):
	
	\begin{itemize}
		\item{Coherence: Evaluates the logical flow and consistency of the plot;}
		\item{Relevance: Evaluates the relevance based on the relationship between the premise and the final script;}
		\item{Interestingness: Evaluates how engaging the script is for readers;}
		\item{Alignment: Evaluates the adherence of the overall trend of plot development with the specified emotional arc type.}
	\end{itemize}
	
	\begin{table}[t]
		\caption{Statistics of human evaluation and machine-human consistency.}
		\centering
		\begin{tabular}{lc}
			\toprule
			\textbf{Metric} & \textbf{Value} \\
			\midrule
			\multicolumn{2}{l}{\textit{Inter-Rater Agreement (Human-Human)}} \\
			ICC & 0.787 \\
			\midrule
			\multicolumn{2}{l}{\textit{Machine-Human Alignment}} \\
			Spearman & 0.656 \\
			\bottomrule
		\end{tabular}
		\label{human_eval}
	\end{table}
	
	To enhance the stability and reliability of the evaluation results, we designed a multiple-scoring mechanism: in a single evaluation, the LLM provides five independent scores and explanations for the same result simultaneously, and the average is taken as the final score, thereby improving the reliability and consistency of the results.
	
	Table \ref{tab:main_results} presents the performance of different methods across the evaluation metrics. Under three different backbone settings, EC-Script significantly outperforms all baseline methods on the Alignment dimension, while maintaining competitive performance on Coherence, Relevance, and Interestingness.
	
	\begin{table*}[t]
		\caption{Qualitative Results via Case Study. This table illustrates key narrative excerpts and corresponding emotional tones, providing a qualitative demonstration of how the generated narrative follows the Cinderella (Rise-Fall-Rise) trajectory.}
		\footnotesize
		\centering
		\renewcommand{\arraystretch}{1.3}
		\begin{tabularx}{\textwidth}{
				>{\raggedright\arraybackslash}p{0.18\textwidth}
				>{\raggedright\arraybackslash}X
				>{\raggedright\arraybackslash}p{0.12\textwidth}
			}
			\toprule
			\textbf{Emotional Phase} & \textbf{Script Excerpt} & \textbf{Emotional Tones} \\
			\midrule
			
			Rise \newline
			\textit{(Scenes 1-3: Jack initially resists Mia's documentary pitch but gradually lowers his defenses. He reconnects with his former bandmate Simon, and despite some awkwardness, begins to open up about his past, finding tentative hope.)} & 
			\textbf{[Scene 3]} \newline
			\textbf{JACK} \newline
			(uneasy, vulnerable) \newline
			Shield's only good until it cracks. After the fallout… I stopped feeling untouchable. Started feeling… exposed. [...] There was a night-after the last show. I wandered off. Ended up in the park, watched the sunrise. Didn't want to go home. Didn't want to face... what I'd done. \newline
			\textbf{MIA} \newline
			(quiet, compassionate) \newline
			\textbf{\textcolor{my_green}{You don't have to carry it alone now.}} \newline
			\textbf{SIMON} \newline
			(old warmth returning) \newline
			Remember the first record? We thought we'd change the world. \textbf{\textcolor{my_green}{Still might, you know-just in a different way.}} \newline
			\textbf{JACK} \newline
			(more open, fragile) \newline
			\textbf{\textcolor{my_green}{If I play, would you sing?}} & 
			Tentative hope, vulnerability, nostalgia, emerging trust, anticipation \\
			
			\midrule
			
			Fall \newline
			\textit{(Scenes 4-8: Jack's attempt to reconnect with his son fails. Soon after, Olivia maliciously leaks unedited, damaging footage. Humiliated and feeling deeply betrayed, Jack retreats into total isolation, hitting an emotional rock bottom.)} & 
			\textbf{[Scene 7]} \newline
			\textbf{SIMON} \newline
			(tense, entering the apartment) \newline
			I saw it. Everybody has. [...] Doesn't matter. It's out, Jack. All of it. \newline
			\textbf{JACK} \newline
			(\textbf{\textcolor{red}{anger flares-quick, desperate}}) \newline
			She said she'd protect it. She— (\textbf{\textcolor{red}{chokes back}}) I trusted her. \newline
			\textbf{[Scene 8]} \newline
			\textbf{JACK} \newline
			(alone in darkness, pacing. He looks at an unopened bottle of whiskey.) \newline
			(bitter laugh) Old friend. Still here. \newline
			(sets it down, returns to the couch, letting the guitar slide to the floor) \newline
			They wanted the truth. Now they've got it. [...] \newline
			(buries his face in his hands) \newline
			\textbf{\textcolor{red}{What's left now? Maybe it's time to disappear.}} & 
			Shock, betrayal, deep humiliation, despair, suffocating isolation \\
			
			\midrule
			
			Rise \newline
			\textit{(Scenes 9-10: With renewed support from Simon and Mia, Jack bravely addresses a live audience. He publicly owns his past mistakes and achieves a heartfelt reconciliation with his son, completing his self-redemption.)} & 
			\textbf{[Scene 10]} \newline
			\textbf{JACK} \newline
			(leans into the mic, voice gaining strength) \newline
			The man you saw in that first cut-drunk, angry, lost-that was me. The man who let down his band, his son, and himself. But that's not the end of the story. I've spent most of my life running from the truth. \textbf{\textcolor{my_green}{But tonight, for once, I want to own it.}} To say I'm sorry-to the people I hurt. To my son. To myself. \textbf{\textcolor{my_green}{I can't rewrite the past, but I can choose what comes next.}} [...] \newline
			\textbf{ETHAN} \newline
			(quiet, from the side of the stage) \newline
			People keep asking if I'm proud of my dad. (shrugs, a half-smile) \textbf{\textcolor{my_green}{Tonight… I think I am.}} \newline
			\textbf{JACK} \newline
			Thank you. (beat) \textbf{\textcolor{my_green}{For letting me begin again.}} & 
			Courage, profound acceptance, reconciliation, authentic joy, triumph \\
			
			\bottomrule
		\end{tabularx}
		\label{narrative_phases}
	\end{table*}

	\subsubsection{Human Evaluation}
	
	To verify the reliability of LLM evaluation, we further conducted human evaluations. We invited six human evaluators to assess the experimental results generated with GPT-4.1 as the generation model. To ensure objectivity, the evaluation was conducted in a blind manner with randomly shuffled outputs from different methods. The evaluation dimensions and criteria were identical to those in the LLM Evaluation. As shown in Table \ref{human_eval}, the scoring among human evaluators achieved reliable consistency (ICC = 0.787), ensuring the validity of the subjective evaluation results. Furthermore, there was a significant positive correlation between human evaluations and automated LLM evaluations (Spearman correlation = 0.656), further confirming the reliability of the LLM evaluation. Due to space constraints and the high consistency between human and LLM evaluation scores, the detailed human evaluation results are omitted. Overall, the human judgments confidently reinforce the automated LLM metrics, confirming that our proposed method effectively adheres to specified affective trajectories while maintaining competitive narrative quality.
	
	\subsubsection{Objective Metrics}
	
	In addition to emotional trajectory adherence evaluation and subjective evaluation, we further conducted objective evaluations of the generated scripts, with specific results shown in the DistinctL-4 and Length columns of Table \ref{tab:main_results}. The Length metric is used to count the average number of words per generated script. Moreover, we use the DistinctL-n\cite{li2025storyteller} metric to evaluate the informativeness and diversity of the generated text. This metric improves upon Distinct-n\cite{li2016diversity} by introducing text length normalization, thereby enabling a fairer evaluation of longer text diversity. Experimental results show that while generating longer scripts, our method achieved good performance on the DistinctL-4 metric, indicating that our narrative effectiveness has genuinely improved, rather than simply increasing text length by padding words.
	
	\begin{equation}
		\label{eq:distinctL}
		\begin{aligned}
			\text{DistinctL} - \text{n} 
			&= \frac{\text{unique n-grams}}{\text{total n-grams}} \\
			&\quad \times \bigl(1 + \log(\text{word\_count})\bigr)
		\end{aligned}
	\end{equation}
	
	\begin{table*}[t]
		\caption{Ablation experiment of individual modules. Bold values indicate the maximum value for that dimension.}
		\centering
		\setlength{\tabcolsep}{4pt}
		\begin{tabular}{cccc >{\columncolor{gray!18}}c}
			\toprule
			\textbf{Method} & 
			\textbf{Coherence$\uparrow$} & 
			\textbf{Relevance$\uparrow$} & 
			\textbf{Interestingness$\uparrow$} & 
			\textbf{Alignment$\uparrow$} \\
			\midrule
			S1           & 60.40 $\pm$ 16.49 & 84.96 $\pm$ 19.17 & 47.55 $\pm$ 10.24 & 80.32 $\pm$ 21.38 \\
			S1 + S2    & \textbf{70.25} $\pm$ 16.92 & 89.22 $\pm$ 13.11 & 59.41 $\pm$ 9.16 & 89.14 $\pm$ 11.07 \\
			S1 + S2 + S3 (Full Model)  & 68.72 $\pm$ 20.67 & \textbf{89.97} $\pm$ 14.67 & \textbf{62.53} $\pm$ 10.36 & \textbf{91.12} $\pm$ 5.76 \\
			\bottomrule
		\end{tabular}
		\label{xiaorong_res}
	\end{table*}

	\subsubsection{Qualitative Result}
	
	To provide a qualitative demonstration of the model's effectiveness in tracking specified affective trajectories, we present an exemplar generated script. Table \ref{narrative_phases} displays key script excerpts and their corresponding emotional tones across different narrative stages for this exemplar.
	
	The premise for this case is: ``An aging rock star faces the consequences of his hedonistic past when a documentary threatens to expose the truth about his life and career". From Scene 1 to Scene 3, protagonist Jack accepts the documentary filming and reconciles with an old friend, gradually stepping out of isolation and remorse as his emotions enter a rising phase. From Scene 4 to Scene 8, documentary footage is maliciously leaked, plunging Jack into a public opinion storm and trust crisis, driving his emotions to rock bottom, corresponding to the falling phase. From Scene 9 to Scene 10, Jack publicly faces his past mistakes, reconciles with his family, and completes his self-redemption, returning to relief and hope, achieving the final rise. The overall emotional development of this case presents a typical ``rise-fall-rise" structure, consistent with the emotional arc type Cinderella, validating the model's capability to follow specified emotional patterns during narrative progression. Such emotional fluctuations help guide individuals to express and release various emotions through narratives, further proving the feasibility and value of our method in emotional healing.
	
	\subsection{Ablation Studies and Analysis}
	
	\begin{figure*}[t]
		\centering
		\includegraphics[width=0.9\linewidth]{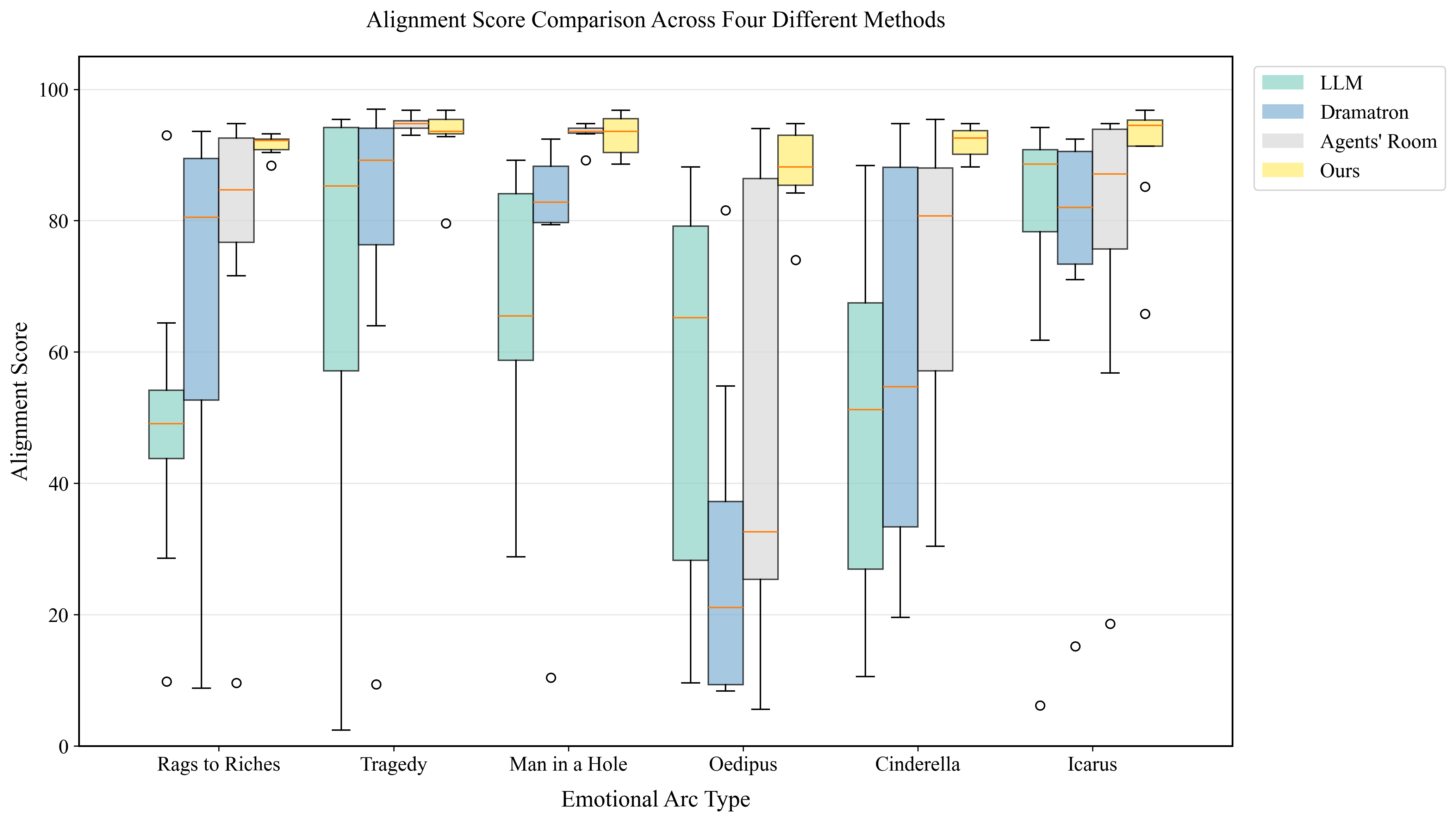}
		\caption{Comparison of alignment scores between EC-Script and baselines across different emotional arc types.}
		\label{boxplot}
	\end{figure*}
	
	\subsubsection{Effectiveness of Individual Modules}
	
	To verify the effectiveness of each functional module within the EC-Script framework, we designed an ablation study by gradually adding modules. Specifically, while maintaining basic script generation capabilities, we sequentially added the Emotion-Trajectory Planning, Character-Driven Scene Generation, and Emotion-Controlled Script Writing modules for comparative experiments. The experimental and evaluation settings were identical to the main experiment.
	
	Ablation settings are as follows:
	
	\begin{itemize}
		\item{S1: Only Emotion-Trajectory Planning; directly generates scripts based on the generated Script Outline;}
		\item{S1 + S2: Adds Character-Driven Scene Generation to S1;}
		\item{S1 + S2 + S3 (Full Model): Our complete method.}
	\end{itemize}
	
	Experimental results are shown in Table \ref{xiaorong_res}. As modules are gradually added, the model's Alignment score continuously improves (80.32 $\rightarrow$ 89.14 $\rightarrow$ 91.12), indicating increasing overall emotion trajectory adherence. The score of 80.32 confirms that Emotion-Trajectory Planning already provides basic structural emotional constraints. After adding Character-Driven Scene Generation, Alignment improves significantly, while Coherence (60.40 $\rightarrow$ 70.25) and Interestingness (47.55 $\rightarrow$ 59.41) also rise synchronously, indicating that character-driven plot planning effectively enhances narrative coherence and expressiveness. With the further addition of Emotion-Controlled Script Writing, although the Alignment increase was smaller, the standard deviation decreased noticeably (11.07 $\rightarrow$ 5.76), showing that this module improves the stability of emotional expression through fine-grained VAD control. At the same time, Interestingness further rose to 62.53, and Relevance maintained stable growth, indicating that local emotional regulation enhances textual expressiveness without disrupting the overall structure. The overall results verify the synergistic role of each module in emotion control and narrative quality, all contributing positively to the final performance.
	
	\subsubsection{Analysis Across Different Emotional Arc Types}
	
	Furthermore, we analyzed the model's performance across different emotional arc types. Figure \ref{boxplot} illustrates the score distribution across different emotional arc types in the Alignment dimension, under the setting of GPT-4.1 as the generation model.
	
	Baseline methods exhibited large fluctuations on complex emotional structures like Cinderella and Oedipus, while showing smaller fluctuations on relatively simpler patterns like Icarus and Man in a Hole, reflecting the differences in generation difficulty among various emotional patterns. In contrast, the score distribution of EC-Script across all types of emotional arcs was much more stable, demonstrating the strong constraining power of our method in emotion control, capable of more robustly and reliably generating narrative content that conforms to specified emotional patterns.
	
	\subsubsection{Computational Cost Analysis}
	
	\begin{table}[htbp]
		\centering
		\caption{Computational cost breakdown of EC-Script modules.}
		\label{tab:cost_analysis}
			\begin{tabular}{ccc}
				\toprule
				\textbf{Module} & \textbf{API Calls} & \textbf{Total Tokens} \\
				\midrule
				S1 & 4 & 3,449 \\
				S2 & 3 & 7,738 \\
				S3 & 17 & 54,520 \\
				Total & 24 & 65,707 \\
				\bottomrule
			\end{tabular}
	\end{table}
	
	To verify feasibility in real-world application scenarios, we quantitatively analyze the computational overhead of the complete EC-Script generation pipeline. As shown in Table \ref{tab:cost_analysis}, generating a full-length drama script with GPT-4.1 as the backbone requires an average of 24 LLM API calls and 65,707 tokens per task. The total inference time varies depending on the selected backbone model, but it generally remains within 10 minutes on average. The majority of the computational investment is concentrated in the scene-by-scene script writing phase (S3), which ensures the precise execution of fine-grained emotional constraints. Given the framework's superior performance in complex emotional control, overall computational cost and inference time remain reasonable and acceptable.
	
	\section{Conclusion}
	
	This paper proposes EC-Script, a long-text script generation method for controllable emotional narratives. Taking emotional arcs as the narrative representation, this method utilizes a hierarchical design to combine Emotion-Trajectory Planning, Character-Driven Scene Generation, and Emotion-Controlled Script Writing based on fine-grained VAD constraints, achieving multi-level emotional modeling from global narrative planning to scene-level text generation. Under experimental settings utilizing several mainstream LLMs as generation backbones, EC-Script significantly outperforms baseline methods in terms of affective trajectory adherence (both Alignment scores and DTW distance). Further ablation studies showed that the various modules operate synergistically to enhance emotion control and narrative quality, validating the effectiveness of the hierarchical emotion modeling strategy. These results demonstrate that structured affective trajectory modeling embedded with a hierarchical control mechanism is critical to improving emotional controllability in complex narrative generation. Furthermore, it provides an effective technical pathway for the research and application of art therapy and emotional healing.
	
	Although EC-Script achieves promising results in emotion controllable long script generation, there are still some limitations that warrant further exploration in the future: (1) While our method provides technical support for emotional healing scenarios, evaluations in real-world therapeutic settings have yet to be conducted to verify its actual therapeutic efficacy. (2) The current pure text generation framework can be extended to multi-modal immersive experiences to enable more diverse emotional healing effects.

	\bibliographystyle{IEEEtran}
	\bibliography{reference}

\end{document}